\documentclass{article}

\usepackage[nonatbib,preprint]{neurips_2022}

\usepackage[utf8]{inputenc} % allow utf-8 input
\usepackage[T1]{fontenc}    % use 8-bit T1 fonts
\usepackage{hyperref} %hyperlinks
\usepackage{url}            % simple URL typesetting
\usepackage{booktabs}       % professional-quality tables
\usepackage{amsfonts}       % blackboard math symbols
\usepackage{nicefrac}       % compact symbols for 1/2, etc.
\usepackage{microtype}      % microtypography
\usepackage{xcolor}         % colors

\usepackage{graphicx}
\usepackage{amsmath}
\usepackage{placeins}

\newcommand{\ie}{\textit{i}.\textit{e}. }
\newcommand{\eg}{\textit{e}.\textit{g}. }

\newcommand\extrafootertext[1]{%
	\bgroup
	\renewcommand\thefootnote{\fnsymbol{footnote}}%
	\renewcommand\thempfootnote{\fnsymbol{mpfootnote}}%
	\footnotetext[0]{#1}%
	\egroup
}

\title{GAUDI: A Neural Architect for \\ Immersive 3D Scene Generation}

\author{ Miguel Angel Bautista$^*$ \And Pengsheng Guo$^*$ \And Samira Abnar \And Walter Talbott \And Alexander Toshev \And Zhuoyuan Chen \And Laurent Dinh \And Shuangfei Zhai \And Hanlin Goh \And Daniel Ulbricht \And Afshin Dehghan \And Josh Susskind \AND

\textnormal{Apple} \\
\url{https://github.com/apple/ml-gaudi}}

\begin{document}

\maketitle

\begin{abstract}
We introduce \textbf{GAUDI}, a generative model capable of capturing the distribution of complex and realistic 3D scenes that can be rendered immersively from a moving camera. We tackle this challenging problem with a scalable yet powerful approach, where we first optimize a latent representation that disentangles radiance fields and camera poses. This latent representation is then used to learn a generative model that enables both unconditional and conditional generation of 3D scenes. Our model generalizes previous works that focus on single objects by removing the assumption that the camera pose distribution can be shared across samples. We show that GAUDI obtains state-of-the-art performance in the unconditional generative setting across multiple datasets and allows for conditional generation of 3D scenes given conditioning variables like sparse image observations or text that describes the scene.
\end{abstract}

\section{Introduction}

\begin{figure}[t]
    \centering
    \includegraphics[trim=105 212 103 215,clip,width=\textwidth]{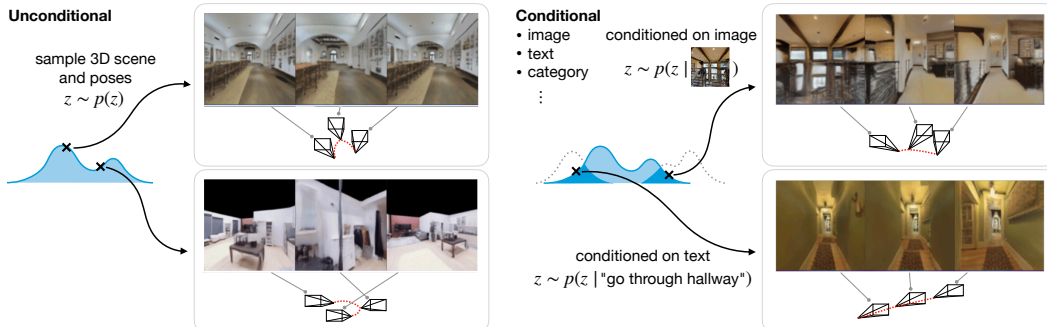} 
    \caption{GAUDI allows to model both conditional and unconditional distributions over complex 3D scenes. Sampled scenes and poses from (left) the unconditional distribution, and (right) a distribution conditioned on an image observation or a text prompt. 
    }
    \label{fig:moneyshot}
\end{figure}

In order for learning systems to be able to understand and create 3D spaces, progress in generative models for 3D is sorely needed. The quote \textit{"The creation continues incessantly through the media of humans."} is often attributed to \textit{Antoni Gaud\'{i}}, who we pay homage to with our method's name. We are interested in generative models that can capture the distribution of 3D scenes and then render views from scenes sampled from the learned distribution. Extensions of such generative models to conditional inference problems could have tremendous impact in a wide range of tasks in machine learning and computer vision. For example, one could sample plausible scene completions that are consistent with an image observation, or a text description (see Fig. \ref{fig:moneyshot} for 3D scenes sampled from GAUDI). In addition, such models would be of great practical use in model-based reinforcement learning and planning \cite{worldmodels}, SLAM \cite{iss}, or 3D content creation. \extrafootertext{$^*$ denotes equal contribution. Corresponding email: \url{mbautistamartin@apple.com}}

Recent works on generative modeling for 3D objects or scenes~\cite{graf, pigan, gsn} employ a Generative Adversarial Network (GAN) where the generator explicitly encodes radiance fields — a parametric function that takes as input the coordinates of a point in 3D space and camera pose, and outputs a density scalar and RGB value for that 3D point. Images can be rendered from the radiance field generated by the model by passing the queried 3D points through the volume rendering equation to project onto any 2D camera view. While compelling on small or simple 3D datasets (\eg single objects or a small number of indoor scenes), GANs suffer from training pathologies including mode collapse~\cite{improvedgan,modecollapse} and are difficult to train on data for which a canonical coordinate system does not exist, as is the case for 3D scenes~\cite{smith2017improved}. In addition, one key difference between modeling distributions of 3D \textit{objects vs. scenes} is that when modeling objects it is often assumed that camera poses are sampled from a distribution that is shared across objects (\ie typically over  $SO(3)$), which is not true for scenes. This is because the distribution of valid camera poses depends on each particular scene independently (based on the structure and location of walls and other objects). In addition, for scenes this distribution can encompass all poses over the $SE(3)$ group. This fact becomes more clear when we think about camera poses as a trajectory through the scene(cf. Fig. \ref{fig:datasets}(b)).

In GAUDI, we map each trajectory (\ie a sequence of posed images from a 3D scene) into a latent representation that encodes a radiance field (\eg the 3D scene) and camera path in a completely disentangled way. We find these latent representations by interpreting them as free parameters and formulating an optimization problem where the latent representation for each trajectory is optimized via a reconstruction objective. This simple training process is scalable to thousands of trajectories. Interpreting the latent representation of each  trajectory as a free parameter also makes it simple to handle a large and variable number of views for each trajectory rather than requiring a sophisticated encoder architecture to pool across a large number of views. After optimizing latent representations for an observed empirical distribution of trajectories, we learn a generative model over the set of latent representations. In the unconditional case, the model can sample radiance fields entirely from the prior distribution learned by the model, allowing it to synthesize scenes by interpolating within the latent space. In the conditional case, conditional variables available to the model at training time (\eg images, text prompts, etc.) can be used to generate radiance fields consistent with those variables. Our contributions can be summarized as:

$\bullet$ We scale 3D scene generation to thousands of indoor scenes containing hundreds of thousands of images, without suffering from mode collapse or canonical orientation issues during training.

$\bullet$ We introduce a novel denoising optimization objective to find latent representations that jointly model a radiance field and the camera poses in a disentangled manner.

$\bullet$ Our approach obtains state-of-the-art generation performance across multiple datasets.

$\bullet$ Our approach allows for various generative setups: unconditional generation as well as conditional on images or text.

\section{Related Work}

In recent years the field has witnessed outstanding progress in generative modeling for the 2D image domain, with most approaches focusing either on adversarial \cite{stylegan, stylegan2} or auto-regressive models \cite{vqvae, vqvae2, vqgan}. More recently, score matching based approaches \cite{scorematching, noneqtherm} have gained popularity. In particular, Denoising Diffusion Probabilistic Models (DDPMs) \cite{ddpm, improvedddpm,ldm,latentscore} have emerged as strong contenders to both adversarial and auto-regressive approaches. In DDPMs, the goal is to learn a step-by-step inversion of a fixed diffusion Markov Chain that gradually transforms an empirical data distribution to a fixed posterior, which typically takes the form of an isotropic Gaussian distribution. In parallel, the last couple of years have seen a revolution in how 3D data is represented within neural networks. By representing a 3D scene as a radiance field, NeRF \cite{nerf} introduces an approach to optimize the weights of a MLP to represent the radiance of 3D points that fall inside the field-of-view of a given set of posed RGB images. Given the radiance for a set of 3D points that lie on a ray shot from a given camera pose, NeRF \cite{nerf} uses volumetric rendering to compute the color for the corresponding pixel and optimizes the MLP weights via a reconstruction loss in image space.

A few attempts have also been made at incorporating a radiance field representation within generative models. Most approaches have focused on the problem of single objects with known canonical orientations like faces or Shapenet objects with shared camera pose distributions across samples in a dataset \cite{graf, pigan, giraffe, nerfvae, eg3d, stylenerf, cips3d, lolnerf}. Extending these approaches from single objects to completely unconstrained 3D scenes is an unsolved problem. One paper worth mentioning in this space is GSN \cite{gsn}, which breaks the radiance field into a grid of local radiance fields that collectively represent a scene. While this decomposition of radiance fields endows the model with high representational capacity, GSN still suffers from the standard training pathologies of GANs, like mode collapse \cite{modecollapse}, which are exacerbated by the fact that unconstrained 3D scenes do not have a canonical orientation. As we show in our experiments (cf. Sect. \ref{sec:experiments}), these issues become prominent as the training set size increases, impacting the capacity of the generative model to capture complex distributions. Separately, a line of recent approaches have also studied the problem of learning generative models of scenes without employing radiance fields \cite{atiss,sceneformer,fastsynth}. These works assume that the model has access to room layouts and a database of object CAD models during training, simplifying the problem of scene generation to a selection of objects from the database and pose predictions for each object.

Finally, approaches that learn to predict a target view given a single (or multiple) source view and relative pose transformation have been recently proposed \cite{videoae, pixelnerf, srt, enr, enrv2}. The pure reconstruction objective employed by these approaches forces them to learn a deterministic conditional function that maps a source image and a relative camera transformation to a target image.  The first is that this scene completion problem is ill-posed (\eg given a single source view of a scene there are multiple target completions that are equally likely). Attempts at modeling the problem in a probabilistic manner have been proposed \cite{geogpt,lookoutsidetheroom}. However, these approaches suffer from inconsistency in predicted scenes because they do not explicitly model a 3D consistent representation like a radiance field.

\section{GAUDI}

Our goal is to learn a generative model given an empirical distribution of trajectories over 3D scenes. Let $X = \{x_{i \in \{0, \dots, n\}}\}$ denote a collection of examples defining an empirical distribution, where each example $x_i$ is a trajectory. Every trajectory $x_i$ is defined as a variable length sequence of corresponding RGB, depth images and $6$DOF camera poses (see Fig. \ref{fig:datasets}).

We decompose the task of learning a generative model in two stages. First, we obtain a latent representation $\mathbf{z}=[\mathbf{z}_\mathrm{scene},\mathbf{z}_\mathrm{pose}]$ for each example $x \in X$ that represents the scene radiance field and pose in separate disentangled vectors. Second, given a set of latents $Z = \{\mathbf{z}_{i \in \{0, \dots, n\}}\}$ we learn the distribution $p(Z)$.

\subsection{Optimizing latent representations for radiance fields and camera poses}

\begin{figure}[t]

    \includegraphics[width=\textwidth]{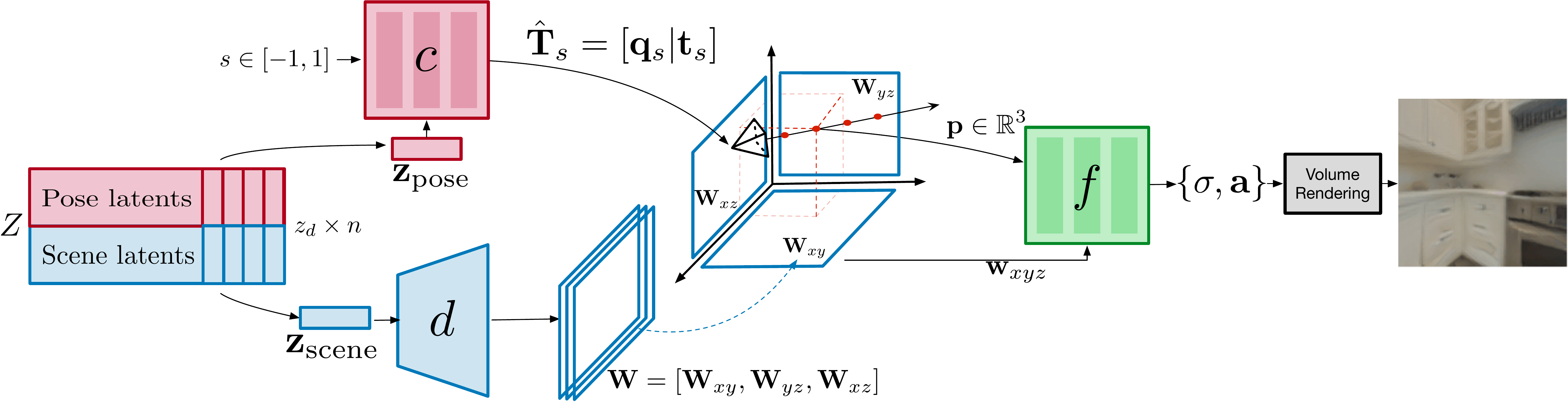}\\
    \caption{Architecture of the decoder model that disentangles camera poses from 3D geometry and appearance of the scene. Our decoder is composed by 3 submodules. A decoder $d$ that takes as input a latent code representing the scene $\mathbf{z}_{\mathrm{scene}}$ and produces a factorized representation of 3D space via a tri-plane latent encoding $\mathbf{W}$. A radiance field network $f$ that takes as input points $\mathbf{p} \in \mathbf{R}^3$ and is conditioned on $\mathbf{W}$ to predict a density $\sigma$ and a signal $\mathbf{a}$ to be rendered via volumetric rendering (Eq. \ref{eq:volumetric_rendering}). Finally, we decode the camera poses through a network $c$ that takes as input a normalized temporal position $s \in [-1, 1]$ and is conditioned on $\mathbf{z}_{\mathrm{pose}}$ which represents camera poses for the whole trajectory $x$ to predict the camera pose $\hat{\mathbf{T}}_s \in SE(3)$.}
    \label{fig:autodecoder}
\end{figure}

We now turn to the task of finding a latent representation $\mathbf{z} \in Z$ for each example $x \in X$ (\ie for each trajectory in the empirical distribution). To obtain this latent representation we take an encoder-less view and interpret $\mathbf{z}$'s as free parameters to be found via an optimization problem \cite{glo, deepsdf}. To map latents $\mathbf{z}$ to trajectories $x$, we design a network architecture (\ie a decoder) that disentangles camera poses and radiance field parameterization. Our decoder architecture is composed of 3 networks (shown in Fig. \ref{fig:autodecoder}):

$\bullet$ The \textbf{camera pose decoder} network $c$ (parameterized by $\theta_c$), is responsible for predicting camera poses $\hat{\mathbf{T}}_s \in SE(3)$ at the normalized temporal position $s \in [-1, 1]$ in the trajectory, conditioned on  $\mathbf{z}_{ \mathrm{pose}}$ which represents the camera poses for the \textit{whole} trajectory. To ensure that the output of $c$ is a valid camera pose (\eg an element of $SE(3)$), we output a $3$D vector representing a \textit{normalized} quaternion $\mathbf{q}_{s}$ for the orientation and a $3$D translation vector $\mathbf{t}_{s}$.

$\bullet$ The \textbf{scene decoder} network $d$ (parameterized by $\theta_d$), is responsible for predicting a conditioning variable for the radiance field network $f$. This network takes as input a latent code that represents the scene $\mathbf{z}_{\mathrm{scene}}$ and predicts an axis-aligned tri-plane representation \cite{convoccnet, eg3d} $\mathbf{W} \in \mathbb{R}^{3 \times S \times S \times F}$. Which correspond to 3 feature maps $[\mathbf{W}_{xy}, \mathbf{W}_{xz}, \mathbf{W}_{yz}]$ of spatial dimension $S \times S$ and $F$ channels, one for each axis aligned plane: $xy$, $xz$ and $yz$.

$\bullet$ The \textbf{radiance field decoder} network $f$ (parameterized by $\theta_f$), is tasked with reconstructing image level targets using the volumetric rendering equation in Eq. \ref{eq:volumetric_rendering}. The input to $f$ is $\mathbf{p} \in \mathbb{R}^3$ and the tri-plane representation $\mathbf{W}=[\mathbf{W}_{xy}, \mathbf{W}_{xz}, \mathbf{W}_{yz}]$. Given a $3$D point $\mathbf{p}=[i, j, k]$ for which radiance is to be predicted, we orthogonally project $\mathbf{p}$ into each plane in $\mathbf{W}$ and perform bi-linear sampling. We concatenate the 3 bi-linearly sampled vectors into $\mathbf{w}_{xyz} = [\mathbf{W}_{xy}(i,j), \mathbf{W}_{xz}(j,k), \mathbf{W}_{yz}(i,k)] \in \mathbb{R}^{3F}$, which is used to condition the radiance field function $f$. We implement $f$ as a MLP that outputs a density value $\sigma$ and a signal $\mathbf{a}$. To predict the value $\mathbf{v}$ of a pixel, the volumetric rendering equation is used (cf. Eq. \ref{eq:volumetric_rendering}) where a 3D point is expressed as ray direction $\mathbf{r}$ (corresponding with the pixel location) at particular depth $u$.
\begin{eqnarray}
\mathbf{v}(\mathbf{r}, \textbf{W}) = \int_{u_n}^{u_f} Tr(u) \sigma\left(\mathbf{r}(u), \mathbf{w}_{xyz}\right) \mathbf{a}\left(\mathbf{r}(u), \mathbf{d}, \mathbf{w}_{xyz}\right) du \nonumber  \\
Tr(u) = \exp \left(-\int_{u_n}^{u} \sigma(\textbf{r}(u), \mathbf{w}_{xyz}) du\right).
\label{eq:volumetric_rendering}
\end{eqnarray}
We formulate a denoising reconstruction objective to jointly optimize for $\theta_d$, $\theta_c$, $\theta_f$ and $\{\mathbf{z}\}_{i=\{0, \dots, n\}}$, shown in Eq. \ref{eq:autodecoder}. Note that while latents $\mathbf{z}$ are optimized for each example $x$ independently, the parameters of the networks $\theta_d$, $\theta_c$, $\theta_f$ are amortized across all examples $x \in X$. As opposed to previous auto-decoding approaches \cite{glo, deepsdf}, each latent $\mathbf{z}$ is perturbed during training with additive noise that is proportional to the empirical standard deviation across all latents, $\mathbf{z} = \mathbf{z} + \beta \mathcal{N}(0, \mathrm{std}(Z))$, inducing a contractive representation \cite{contractive}. In this setting, $\beta$ controls the trade-off between the entropy of the distribution $\mathbf{z} \in Z$ and the reconstruction term, with $\beta=0$ the distribution of $\mathbf{z}$'s becomes a collection of indicator functions, whereas non-trivial structure in latent space arises for $\beta > 0$. We use a small $\beta > 0 $ value to enforce a latent space in which interpolated samples (or samples that contain small deviations from the empirical distribution, as the ones that one might get from sampling a subsequent generative model) are included in the support of the decoder.
\begin{equation}
    \min_{\theta_d, \theta_f, \theta_c, Z} \mathbb{E}_{x \sim X} \left[\mathcal{L}_{\mathrm{scene}}(\mathbf{x}^{\mathrm{im}}_{s}, \mathbf{z}_{\mathrm{scene}}, \mathbf{T}_{s} )  + \lambda \mathcal{L}_{\mathrm{pose}}(\mathbf{T}_{s}, \mathbf{z}_{ \mathrm{pose}}, s) \right]
    \label{eq:autodecoder}
\end{equation}
We optimize parameters $\theta_d, \theta_f, \theta_c$ and latents $\mathbf{z} \in Z$ with two different losses. The first loss function $\mathcal{L}_{\mathrm{scene}}$ measures the reconstruction between the radiance field encoded in $\mathbf{z}_{\mathrm{scene}}$ and the images in the trajectory $\mathbf{x}^{\mathrm{im}}_{s}$ (where $s$ denotes the normalized temporal position of the frame in the trajectory), given ground-truth camera poses $\mathbf{T}_{s}$ required for rendering. We use an $l_2$ loss for RGB and $l_1$ for depth \footnote{We obtain depth predictions by aggregating densities across a ray as in \cite{nerf}}. The second loss function $\mathcal{L}_{\mathrm{pose}}$ measures the camera pose reconstruction error between the poses  $\hat{\mathbf{T}}_{s}$ encoded in $\mathbf{z}_{\mathrm{pose}}$ and the ground-truth poses. We employ an $l_2$ loss on translation and $l_1$ loss for the normalized quaternion part of the camera pose. Although theoretically normalized quaternions are not necessarily unique (\eg $\mathbf{q}$ and $- \mathbf{q}$) we do not observe any issues empirically during training.

\subsection{Prior Learning}

Given a set of latents $\mathbf{z} \in Z$ resulting from minimizing the objective in Eq. \ref{eq:autodecoder}, our goal is to learn a generative model $p(Z)$ that captures their distribution (\ie after minimizing the objective in Eq. \ref{eq:autodecoder} we interpret $\mathbf{z} \in Z$ as examples from an empirical distribution in latent space). In order to model $p(Z)$ we employ a Denoising Diffusion Probabilistic Model (DDPM) \cite{ddpm}, a recent score-matching \cite{scorematching} based model that learns to reverse a diffusion Markov Chain with a large but finite number of timesteps. In DDPMs \cite{ddpm} it is shown that this reverse process is equivalent to learning a sequence of denoising auto-encoders with tied weights. The supervised denoising objective in DDPMs makes learning $p(Z)$ simple and scalable. This allows us to learn a powerful generative model that enables both unconditional and conditional generation of 3D scenes. For training our prior $p_{\theta_p}(Z)$ we take the objective function in \cite{ddpm} defined in Eq. \ref{eq:ddpm_simple}.  In Eq. \ref{eq:ddpm_simple} $t$ denotes the timestep, $\epsilon \sim \mathcal{N}(0, \mathbf{I})$ is the noise and $\bar{\alpha}_t$ is a noise magnitude parameter with a fixed scheduling. Finally, $\epsilon_{\theta_p}$ denotes the denoising model.

\begin{equation}
\min_{\theta_p} \mathbb{E}_{t, \mathbf{z} \sim Z, \epsilon \sim \mathcal{N}(0, \mathbf{I})} \left[\| \epsilon - \epsilon_{\theta_p}( \sqrt{\bar{\alpha}_t} \mathbf{z}  + \sqrt{1 - \bar{\alpha}_t}\epsilon , t) \|^2 \right]
\label{eq:ddpm_simple}
\end{equation}
At inference time, we sample $\mathbf{z} \sim p_{\theta_p}(Z)$ by following the inference process in DDPMs. We start by sampling $\mathbf{z}_T \sim \mathcal{N}(0, \mathbf{I})$ and iteratively apply $\epsilon_{\theta_p}$ to gradually denoise $\mathbf{z}_T$, thus reversing the diffusion Markov Chain to obtain $\mathbf{z}_0$. We then feed $\mathbf{z}_0$ as input to the decoder architecture (cf. Fig. \ref{fig:autodecoder}) and reconstruct a radiance field and a camera path.

If the goal is to learn a conditional distribution of the latents $p(Z | Y)$, given paired data $\{\mathbf{z} \in Z, y \in Y\}$, the denoising model $\epsilon_\theta$ is augmented with a conditioning variable $y$, resulting in $\epsilon_{\theta_p}(\mathbf{z}, t, y)$, implementation details about how the conditioning variable is used in the denoising architecture can be found in the appendix \ref{app:architecture}. 

\section{Experiments}
\label{sec:experiments}

In this section we show the applicability of GAUDI to multiple problems. First, we evaluate reconstruction quality and performance of the reconstruction stage. Then, we evaluate the performance of our model in generative tasks including unconditional and conditional inference, in which radiance fields are generated from conditioning variables corresponding to  images or text prompts. Full experimental settings and details can be found in the appendix \ref{app:expdetails}.

\subsection{Data}
\begin{figure}
    \centering

    \includegraphics[width=\textwidth]{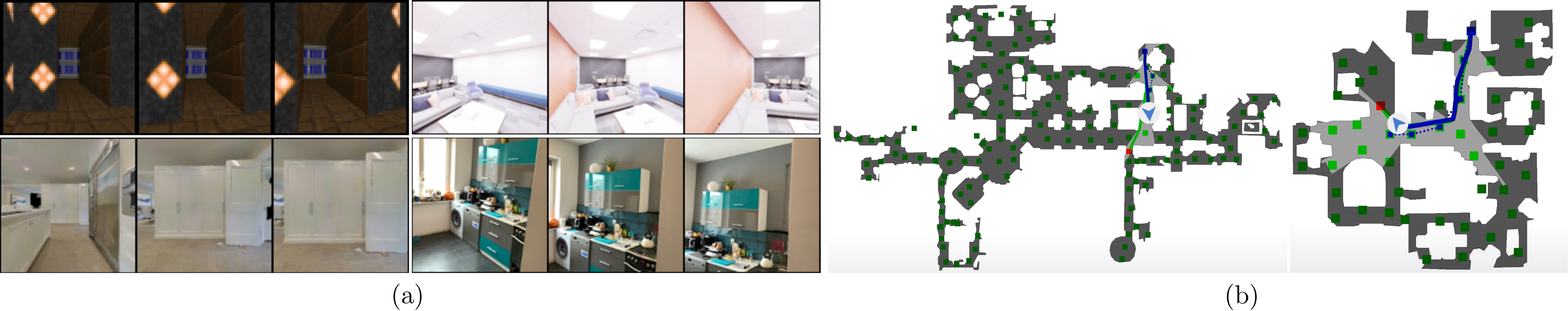}
    \caption{(a) Examples of the 4 datasets we use in this paper (from left to right): Vizdoom \cite{vizdoom}, Replica \cite{replica}, VLN-CE \cite{vlnce}, ARKitScenes \cite{arkit}. (b) Top-down views of 2 different camera paths in VLN-CE \cite{vlnce}. Blue and red dots represent start-end positions and the camera path is highlighted in blue.}
    \label{fig:datasets}
\end{figure}

We report results on 4 datasets: Vizdoom \cite{vizdoom}, Replica \cite{replica}, VLN-CE \cite{vlnce} and ARKit Scenes \cite{arkit}, which vary in number of scenes and complexity (see Fig. \ref{fig:datasets} and Tab. \ref{tab:reconstruction}). 

\textbf{Vizdoom} \cite{vizdoom}: Vizdoom is a synthetic simulated environment with simple texture and geometry. We use the data provided by \cite{gsn} to train our model. It is the simplest dataset in terms of number of scenes and trajectories, as well as texture, serving as a test bed to examine GAUDI in the simplest setting.

\textbf{Replica} \cite{replica}: Replica is a dataset comprised of $18$ realistic scenes from which trajectories are rendered via Habitat \cite{habitat}. We used the data provided by \cite{gsn} to train our model.

\textbf{VLN-CE} \cite{vlnce}: VLN-CE is a dataset originally designed for vision and language navigation in continuous environments. This dataset is composed of $3.6$K trajectories of an agent navigating between two points in a 3D scene from the 3D dataset \cite{chang2017matterport3d}. We render observations via Habitat \cite{habitat}. Notably, this dataset contains also textual descriptions of the trajectories taken by an agent. In Sect. \ref{sec:conditional_experiments} we train GAUDI in a conditional manner to generate 3D scenes given a description.

\textbf{ARKitScenes} \cite{arkit}: ARKitScenes is a dataset of scans of indoor spaces. This dataset contains more than $5$K scans of about $1.6$K different indoor spaces. As opposed to the previous datasets where RGB, depth and camera poses are obtained via rendering in a simulation (\ie either Vizdoom \cite{vizdoom} or Habitat \cite{habitat}), ARKitScenes provides raw RGB and depth of the scans and camera poses estimated using ARKit SLAM. In addition, whereas trajectories from the previous datasets are point-to-point, as typically done in navigation, the camera trajectories for ARKitScenes resembles a natural scan a of full indoor space. In our experiments we use a subset of $1$K scans from ARKitScenes to train our models.

\subsection{Reconstruction}
\label{sec:reconstruction}

\begin{figure}[h]
    \centering

    \includegraphics[width=\textwidth]{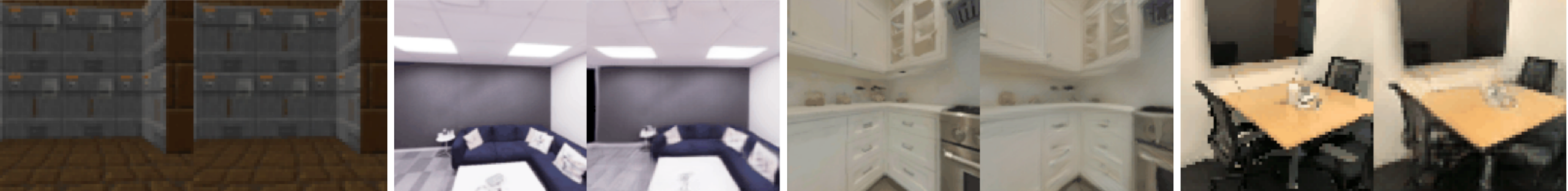}
    \caption{Qualitative reconstruction results of random trajectories on different datasets (one for each column): Vizdoom \cite{vizdoom}, Replica \cite{replica}, VLN-CE \cite{vlnce}and ARKitScenes \cite{arkit}. For each pair of images the left is ground-truth and right is reconstruction.}
    \label{fig:reconstruction}
\end{figure}

We first validate the hypothesis that the optimization problem described in Eq. \ref{eq:autodecoder} can find latent codes $\mathbf{z}$ that are able reconstruct the trajectories in the empirical distribution in a satisfactory way. In Tab. \ref{tab:reconstruction} we report reconstruction performance of our model across all datasets. Fig. \ref{fig:reconstruction} shows reconstructions of random trajectories for each dataset. For all our experiments we set the dimension of $\mathbf{z}_{\mathrm{scene}}$ and $ \mathbf{z}_{\mathrm{pose}}$ to 2048 and $\beta=0.1$ unless otherwise stated. During training, we normalize camera poses for each trajectory so that the middle frame in a trajectory becomes the origin of the coordinate system. See appendix \ref{app:ablation} for ablation experiments.

\begin{table}[h]
\tiny
    \begin{center}
    \begin{tabular}{ c c | c c  c  |c c }
    \toprule
            & \#sc-\#tr-\#im & $l_1$ $\downarrow$ & PSNR $\uparrow$ & SSIM $\uparrow$ & Rot Err. $\downarrow$ & Trans. Err $\downarrow$ \\ 
    \midrule
          Vizdoom \cite{vizdoom} & 1-32-1k & 0.004 & 44.42 & 0.98  & 0.01 & 1.26 \\ 
          Replica  \cite{replica} & 18-100-1k  & 0.006 & 38.86 & 0.99 & 0.03 & 0.01 \\ 
          VLN-CE \cite{vlnce} & 90-3.6k-600k  & 0.031  & 25.17 & 0.73  & 0.30 & 0.02 \\ 
          ARKitScenes \cite{arkit} &  300-1k-600k  & 0.039 & 24.51 & 0.76  & 0.16 & 0.04\\ 
    \bottomrule
    \end{tabular}
    \end{center}
    \caption{Reconstruction results of the optimization process described in Eq. \ref{eq:autodecoder}. The first column shows the number of scenes (\#sc), trajectories (\#tr) and images (\#im) per dataset. Due to the large number of images on VLN-CE \cite{vlnce} and ARKitScenes \cite{arkit} datasets we sample 10 random images per trajectory to compute the reconstruction metrics. }
    \label{tab:reconstruction}
\end{table}

\subsection{Interpolation}

\begin{figure}[h]
    \centering
    \includegraphics[width=\textwidth]{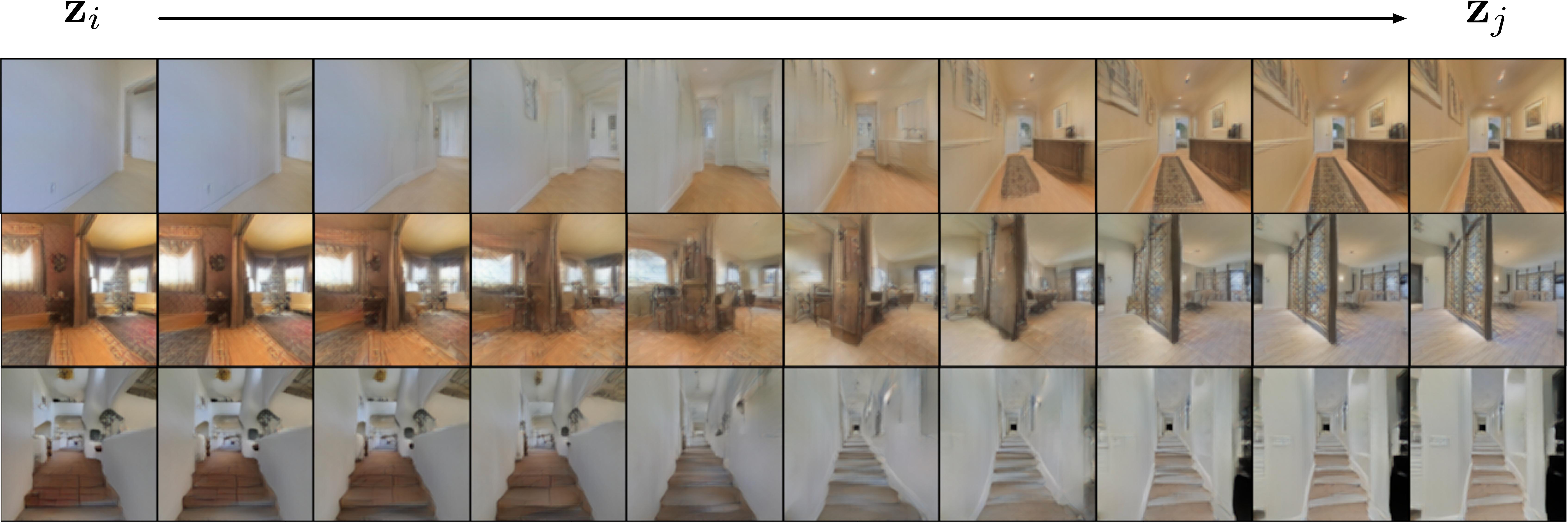} 
    \caption{Interpolation of 3D scenes in latent space (\eg interpolating the encoded radiance field) for the VLN-CE dataset \cite{vlnce}. Each row corresponds to a different interpolation path.}
    \label{fig:interpolation}
\end{figure}

In addition, to evaluate the structure of the latent representation obtained from minimizing the optimization problem in Eq. \ref{eq:autodecoder}, we show interpolation results between pairs of latents  $(\mathbf{z}_i, \mathbf{z}_j)$ in Fig. \ref{fig:interpolation}. To render images while interpolating the scene we place a fixed camera at the origin of the coordinate system. We observe a smooth transition of scenes in both geometry (walls, ceilings) and texture (stairs, carpets). More visualizations are included in the appendix \ref{app:visualizations}. 

\subsection{Unconditional generative modeling}

Given latent representations $\mathbf{z} \in Z$ that can reconstruct samples $x \in X$ with high accuracy as shown in Sect. \ref{sec:reconstruction}, we now evaluate the capacity of the prior $p_{\theta_p}(Z)$ to capture the empirical distribution $x \in \mathcal{X}$ by learning the distribution of latents $\mathbf{z}_i \in Z$. To do so we sample $\mathbf{z} \sim p_{\theta_p}(Z)$ by following the inference process in DDPMs, and then feed $\mathbf{z}$ through the decoder network, which results in trajectories of RGB images that are then used for evaluation. We compare our approach with the following baselines: GRAF \cite{graf}, $\pi$-GAN \cite{pigan} and GSN \cite{gsn}. We sample $5$k images from predicted and target distributions for each model and dataset and report both FID \cite{fid} and SwAV-FID \cite{swavfid} scores. We report quantitative results in Tab. \ref{tab:generative_quant}, where we can see that GAUDI obtains state-of-the-art performance across all datasets and metrics. We attribute this performance improvement to the fact that GAUDI learns disentangled yet corresponding latents for radiance fields and camera poses, which is key when modeling scenes (see ablations in the appendix \ref{app:ablation}). We note that to obtain these great empirical results GAUDI needs to simultaneously find latents with high reconstruction fidelity while also efficiently learning their distribution.

\begin{table}[h]
\tiny
\begin{center}
\resizebox{\columnwidth}{!}{%
\setlength\tabcolsep{2pt} % default value: 6pt
 \begin{tabular}{lcccccccccc}
    \toprule
    &  \multicolumn{2}{c}{VizDoom \cite{vizdoom}} & \multicolumn{2}{c} {Replica \cite{replica}} & \multicolumn{2}{c} {VLN-CE \cite{vlnce}} & \multicolumn{2}{c} {ARKitScenes \cite{arkit}} \\
   \cmidrule(r){2-3}\cmidrule(r){4-5}\cmidrule(r){6-7} \cmidrule(r){8-9}
  & FID $\downarrow$ & SwAV-FID $\downarrow$ & FID $\downarrow$ & SwAV-FID $\downarrow$ & FID $\downarrow$& SwAV-FID $\downarrow$ & FID $\downarrow$& SwAV-FID $\downarrow$   \\ 
 \midrule 
  GRAF \cite{graf} &  $47.50 \pm 2.13$ & $5.44 \pm 0.43$ & $65.37 \pm 1.64$ & $5.76 \pm 0.14$  & $90.43 \pm 4.83$ & $8.65 \pm 0.27$ & $87.06 \pm 9.99$  & $13.44 \pm 0.26$\\
  $\pi$-GAN\cite{pigan} &  $143.55 \pm 4.81$  & $15.26 \pm 0.15$ & $166.55 \pm 3.61$ & $13.17 \pm 0.20$ &  $151.26 \pm 4.19674$ & $14.07 \pm 0.56$ & $134.80 \pm 10.60$ & $ 15.58 \pm 0.13$ \\
  GSN \cite{gsn} & $37.21 \pm 1.17$ & $4.56 \pm 0.19$  & $41.75 \pm 1.33$ & $4.14 \pm 0.02$ & $43.32 \pm 8.86$ & $6.19 \pm 0.49$ & $79.54 \pm 2.60$ & $10.21 \pm 0.15$\\
  GAUDI &  $\mathbf{33.70 \pm 1.27}$ &  $\mathbf{3.24 \pm 0.12}$ & $\mathbf{18.75 \pm 0.63}$ & $\mathbf{1.76 \pm 0.05}$  &  $\mathbf{18.52\pm 0.11}$ & $\mathbf{3.63\pm 0.65}$ & $\mathbf{37.35\pm 0.38}$ & $\mathbf{4.14\pm 0.03}$\\
  \bottomrule
\end{tabular} %
}
\end{center}
\caption{Generative performance of state-of-the-art approaches for generative modelling of radiance fields on 4 scene datasets: Vizdoom \cite{vizdoom}, Replica \cite{replica}, VLN-CE \cite{vlnce} and ARKitScenes \cite{arkit}, according to FID~\cite{fid} and SwAV-FID \cite{swavfid} metrics.}
\label{tab:generative_quant}
\end{table}

\begin{figure}[h]
    \centering

    \includegraphics[width=\textwidth]{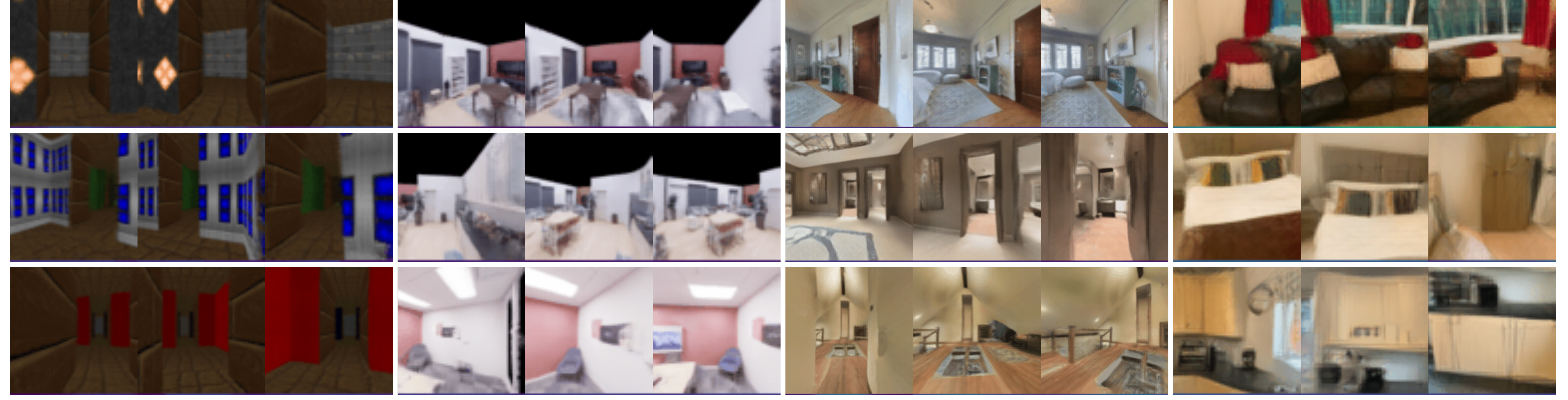}
    \caption{Different scenes sampled from unconditional GAUDI (one sample per row) and rendered from their corresponding sampled camera poses (one dataset per column): Vizdoom \cite{vizdoom}, Replica \cite{replica}, VLN-CE \cite{vlnce}and ARKitScenes \cite{arkit}. The resolutions are $64 \times 64$, $64 \times 64$, $128 \times 128$ and $64 \times 64$ respectively.}
    \label{fig:unconditional_samples}
\end{figure}

In Fig. \ref{fig:unconditional_samples} we show samples from the unconditional distribution learnt by GAUDI for different datasets. We observe that GAUDI is able to generate diverse and realistic 3D scenes from the empirical distribution which can be rendered from the sampled camera poses.

\subsection{Conditional Generative Modeling} 
\label{sec:conditional_experiments}

In addition to modeling the distribution $p(Z)$, with GAUDI we can also tackle conditional generative problems $p(Z|Y)$, where a conditioning variable $y \in Y$ is given to modulate $p(Z)$. For all conditioning variables $y$ we assume the existence of paired data $\{ \mathbf{z}, y \}$ to train the conditional model \cite{vqvae2, vqgan, dalle}. In this section we show both quantitative and qualitative results for conditional inference problems. The first conditioning variable we consider are textual descriptions of trajectories. Second, we consider a conditional model where randomly sampled RGB images in a trajectory act as conditioning variables. Finally, we use a categorical variable that indicates the 3D environment (\ie the particular indoor space) from which each trajectory was obtained (\ie a one-hot vector). Tab. \ref{tab:cond_generative_quant} shows quantitative results for the different conditional inference problems.

\begin{table}[h]
\tiny
\begin{center}
\begin{tabular}{lcccccccc}

    \toprule
     \multicolumn{2}{c}{Text Conditioning} & \multicolumn{2}{c} {Image Conditioning} & \multicolumn{2}{c} {Categorical Conditioning} \\
    \cmidrule(r){1-2}\cmidrule(r){3-4}\cmidrule(r){5-6}\cmidrule(r){7-8}
    FID $\downarrow$ & SwAV-FID $\downarrow$ & FID $\downarrow$ & SwAV-FID $\downarrow$ & FID $\downarrow$& SwAV-FID $\downarrow$ \\ 
    \midrule 
    $ 18.50 $ & $ 3.75 $  &  $ 19.51 $ & $ 3.93 $ & $ 18.74 $ &  $ 3.61 $ \\
    \bottomrule
\end{tabular}
\quad\quad\quad
\begin{tabular}{lcccc}

    \toprule
     \multicolumn{2}{c}{Avg. $ \Delta$ Per-Environment} \\
    \cmidrule(r){1-2}
    FID $\downarrow$ & SwAV-FID $\downarrow$  \\ 
    \midrule 
    $-50.79 $ &  $ -4.10 $\\
    \bottomrule
\end{tabular}
\end{center}
\caption{Quantitative results of Conditional Generative Modeling on VLN-CE \cite{vlnce} dataset. GAUDI is able to produce high-quality scene renderings with low FID and SwAV-FID scores. In the right table we show the difference in average \textit{per-environment} FID score between the conditional and unconditional models.}
\label{tab:cond_generative_quant}
\end{table}
\subsubsection{Text Conditioning}
\begin{figure}[h]
    \centering

    \includegraphics[width=\textwidth]{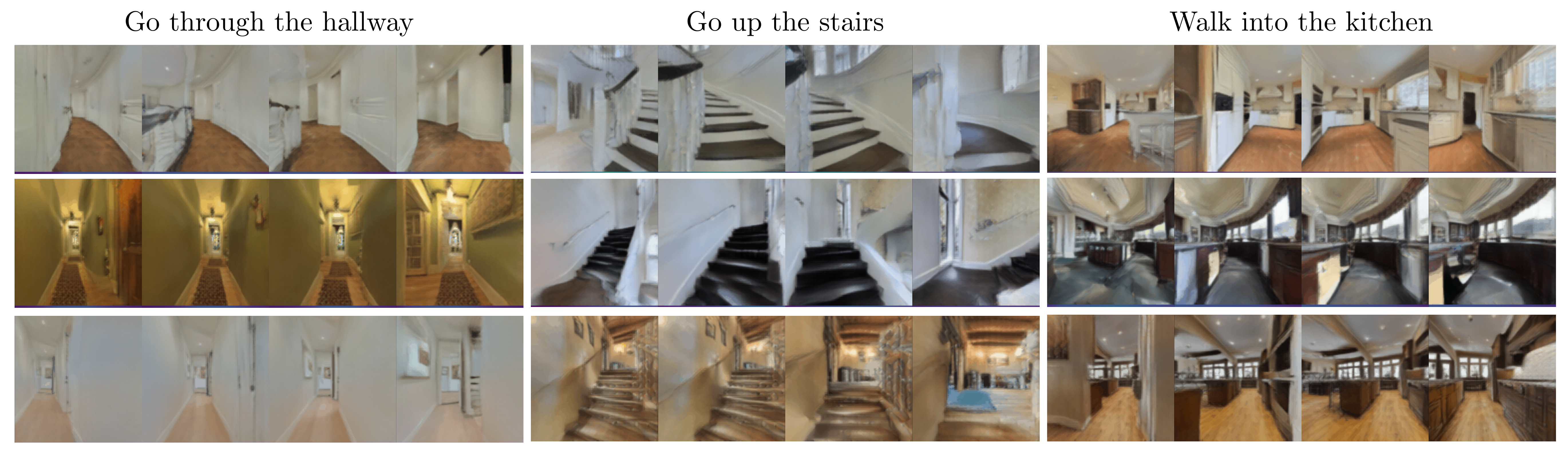}
    \caption{Text conditional 3D scene generation using GAUDI (one sample per row). Our model is able to capture the conditional distributions of scenes by generating multiple plausible scenes and camera paths that match the given text prompts.}
    \label{fig:text_cond}
\end{figure}

We tackle the challenging task of training a text conditional model for 3D scene generation. We use the navigation text descriptions provided in VLN-CE \cite{vlnce} to condition our model. These text descriptions contain high level information about the scene as well as the navigation path (\ie \textit{"Walk out of the bedroom and into the living room"}, \textit{"Exit the room through the swinging doors and then enter the bedroom"}). We employ a pre-trained RoBERTa-base \cite{roberta} text encoder and use its intermediate representation to condition the diffusion model. Fig. \ref{fig:text_cond} shows qualitative results of GAUDI for this task. To the best of our knowledge, this is the first model that allows for conditional 3D scene generation from text in an amortized manner (\ie without distilling CLIP \cite{clip} through a costly optimization problem \cite{dreamfields, text2mesh}).

\subsubsection{Image Conditioning}

\begin{figure}[h]
    \centering

    \includegraphics[width=\textwidth]{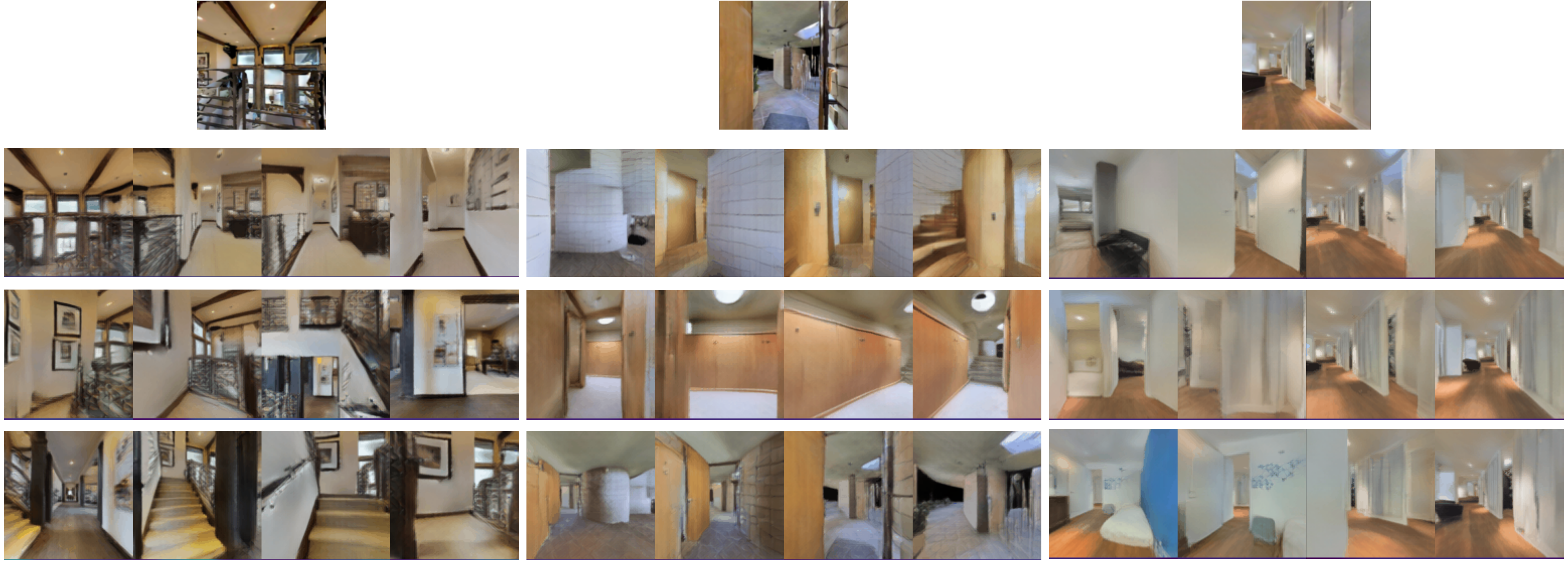}
    \caption{Image conditional 3D scene generation using GAUDI (one sample per row). Given a conditioned image (top row), our model is able to sample scenes where the same or contextually similar view is observed from a stochastic viewpoint.}
    \label{fig:image_cond}
\end{figure}

We now analyze whether GAUDI is able to pick up information from the RGB images to predict a distribution over $Z$. In this experiment we randomly pick images in a trajectory $x \in X$ and use it as a conditioning variable $y$. For this experiment we use trajectories in the VLN-CE dataset \cite{vlnce}. During each training iteration we sample a random image for each trajectory $x$ and use it as a conditioning variable. We employ a pre-trained ResNet-18 \cite{resnet} as an image encoder. During inference, the resulting conditional GAUDI model is able to sample radiance fields where the given image is observed from a stochastic viewpoint. In Fig. \ref{fig:image_cond} we show samples from the model conditioned on different RGB images.

\begin{figure}[h]
    \centering

    \includegraphics[width=\textwidth]{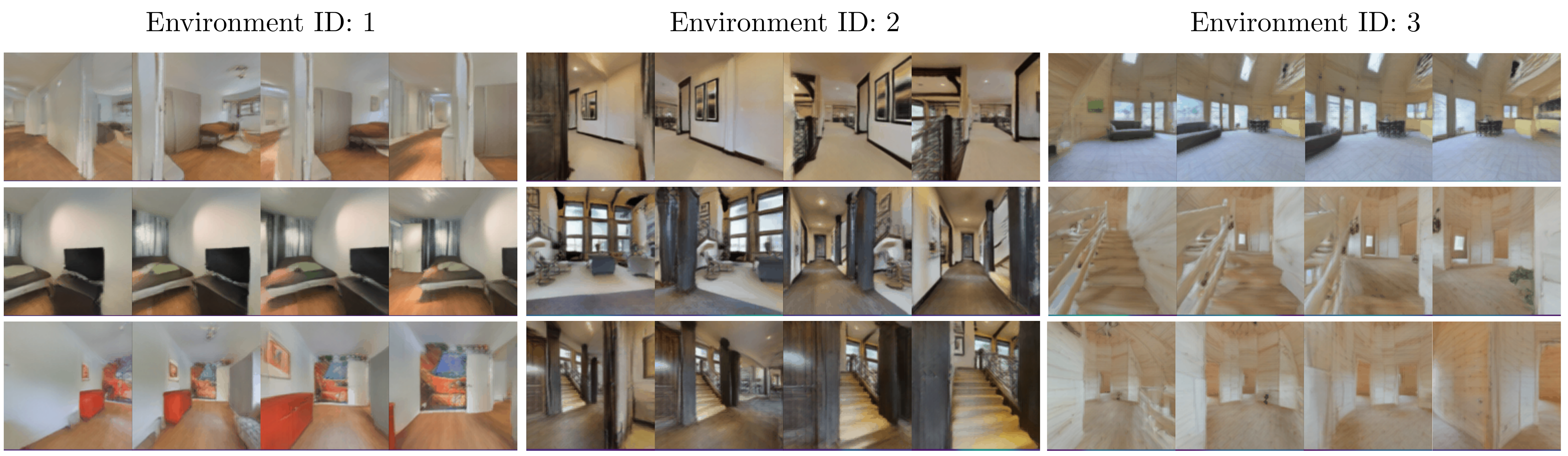}
    \caption{Samples from the GAUDI model conditioned on a categorical variable denoting the indoor scene (one sample per row).}
    \label{fig:scene_id_qual}
\end{figure}

\subsubsection{Categorical Conditioning}

Finally, we analyze how GAUDI performs when conditioned on a categorical variable that indicates the underlying 3D indoor environment in which each trajectory was recorded. We perform experiments in the VLN-CE \cite{vlnce} dataset, where we employ a trainable embedding layer to learn a representation for categorical variables indicating each environment. We compare the \textit{per-environment} FID score of conditional model with its unconditional counterpart. This \textit{per-enviroment} FID score is computed only on real images of the same indoor environment that the model is conditioned on. Our hypothesis is that if the model efficiently captures the information in the conditioning variable it should capture the environment specific distribution better than its unconditional counterpart trained on the same data. In Tab. \ref{tab:cond_generative_quant} the last column shows difference (\eg the $\Delta$) on the average \textit{per-environment} FID score between the conditional and unconditional model on VLN-CE dataset. We observe that the conditional model consistently obtains a better FID score than the unconditional model across all indoor environments, resulting in a sharp reduction of average FID and SwAV-FID scores.  In addition, in Fig. \ref{fig:scene_id_qual} we show samples from the model conditioned on a given categorical variable.

\section{Conclusion}

We have introduced GAUDI, a generative model that captures distributions of complex and realistic 3D scenes. GAUDI uses a scalable two-stage approach which first involves learning a latent representation that disentangles radiance fields and camera poses. The distribution of disentangled latent representations is then modeled with a powerful prior. Our model obtains state-of-the-art performance when compared with recent baselines across multiple 3D datasets and metrics. GAUDI can be used both for conditional and unconditional problems, and enabling new tasks like generating 3D scenes from text descriptions.

{\small
\bibliographystyle{plain}
%\bibliography{bib.bib}

\newpage
\appendix

\section{Limitations, Future Work and Societal Impact}
~\label{app:futurework}

Although GAUDI represents a step forward in generative models for 3D scenes, we would like to clearly discuss the limitations. One current limitation of our model is the fact that inference is not real-time. The reason for this is two fold: (i) sampling from the DDPM prior is slow even if it is amortized for the whole 3D scene. Techniques for improving inference efficiency in DDPMs have been recently proposed \cite{ddim, fastddpm1, fastddpm2} and can complement GAUDI. (ii) Rendering from a radiance field is not as efficient as rendering other 3D structures like meshes. Recent work have also tackled this problem \cite{nsvf, plenoxels, kilonerf} and could be applied to our approach. In addition, many of the latest image generative models \cite{dalle, glide, imagen} use multiple stages of up-sampling through diffusion models to render high-res images. These up-sample stages could be directly applied to GAUDI. In addition, one could considering studying efficient encoders to replace the optimization process to find latents. While attempts have been made at using transformers \cite{srt} for short trajectories (5-10 frames) it is unclear how to scale to thousands of images per trajectory like the ones in \cite{arkit}. Finally, the main limitation for a model like GAUDI to exhibit improved generation and generalization abilities is the lack of massive-scale and open-domain 3D datasets. In particular ones with other associated modalities like textual descriptions. 

When considering societal impact of generative models a few aspects that need attention are the use generative models for creating disingenuous data, \eg "DeepFakes" \cite{deepfakes}, training data leakage and privacy \cite{leakage}, and amplification of the biases present in training data \cite{amplification}. One specific ethical consideration that applies to GAUDI is the impact that a model which can easily create immersive 3D scenes can have on future generations and their detachment of reality \cite{ethicmr}. For an in-depth review of ethical considerations in generative modeling we refer the reader to \cite{ethics}.

\section{Experimental Settings and Details}
~\label{app:expdetails}

In this section we describe details about data and model hyper-parameters. For all experiments our latents $\mathbf{z}_{\mathrm{scene}}$ and $\mathbf{z}_{\mathrm{pose}}$ have $2048$ dimensions. In the first stage, when latents are optimized via Eq. \ref{eq:autodecoder}, $\mathbf{z}_{\mathrm{scene}}$ gets reshaped to a $8 \times 8 \times 32$ feature map before feeding it to the scene decoder network. In the second stage, when training the DDPM prior we reshape $\mathbf{z}_{\mathrm{scene}}$ and $\mathbf{z}_{\mathrm{pose}}$ to $8 \times 8 \times 64$ latent and leverage the power of a UNet \cite{unet} denoising architecture.

For each dataset, trajectories have different length, physical scale, as well as near and far planes for rendering, which we adjust accordingly in our model.

\textbf{Vizdoom} \cite{vizdoom}: In Vizdoom, trajectories contains $600$ steps on average. In each step the camera is allowed to move forward $0.5$ game units or rotate left or right by $30$ degrees. We set the unit length of an element in the tri-plane representation as $0.05$ game units (meaning each latent code $\mathbf{w}_{xyz}$ represents a volume of space of $0.05$ cubic game units). The near plane is at $0.0$ game units and the far plane at $800$ game units. We use the data and splits provided by \cite{gsn}.

\textbf{Replica} \cite{replica}: In Replica, all trajectories contain 100 steps. In each step, the camera can either rotate left or right by $25$ degrees or move forward $15$ centimeters. We set the unit length of an element in the tri-plane representation as $25$ centimeters (meaning each latent code $\mathbf{w}_{xyz}$ represents a volume of space of $0.25$ cubic centimeters). The near plane is at $0.0$ meters and the far plane at $6$ meters. We use the data and splits provided by \cite{gsn}.

\textbf{VLN-CE} \cite{vlnce}: in VLN-CE trajectories contain a variable number of steps between $30$ and $150$, approximately. In each step, the camera can either rotate left or right by $25$ degrees or move forward $15$ centimeters. We set the unit length of an element in the tri-plane representation as $50$ centimeters. The near plane is at $0.0$ meters and the far plane at $12$ meters. We use the data and training splits provided by \cite{vlnce}.

\textbf{ARKitScenes} \cite{arkit}: in ARKitScenes trajectories contain a number of steps around $1000$ on average. In these trajectories the camera is able to move continuously in any direction and orientation. We set the unit length of an element in the tri-plane representation as $20$ centimeters. The near plane is at $0.0$ meters and the far plane at $8$ meters. We use the 3DOD split of data provided by \cite{arkit}

\section{Decoder Architecture Design and Details}
~\label{app:architecture}
In this section we describe the decoder model in Fig. \ref{fig:autodecoder} in the main paper. The decoder network is composed of 3 modules: \textbf{scene decoder}, \textbf{camera pose decoder} and \textbf{radiance field decoder}. 

\begin{itemize}
    \item The \textbf{scene decoder} network follows the architecture of the VQGAN decoder \cite{vqgan}, parameterized with convolutional architecture that contains a self-attention layers at the end of each block. The output of the scene decoder is a feature map of shape $64 \times 64 \times 768$. To obtain the tri-plane representation $\mathbf{W} = [\mathbf{W}_{xy}, \mathbf{W}_{xz}, \mathbf{W}_{yz}]$ we split the channel dimension of the output feature map in 3 chunks of equal size $64 \times 64 \times 256$.
    
    \item The \textbf{camera pose decoder} is implemented as an MLP with $4$ conditional batch normalization (CBN) blocks with residual connections and hidden size of $256$, as in \cite{occnet}. The conditional batch normalization parameters are predicted from $\mathbf{z}_{\mathrm{pose}}$. We apply positional encoding to the inputs the camera pose encoder ($s \in [-1, 1]$).  Fig. \ref{fig:model_architectures}(a) shows the architecture of the camera pose decoder module.

    \item The \textbf{radiance field decoder} is implemented as an MLP with $8$ linear layers with hidden dimension of $512$ and LeakyReLU activations. We apply positional encoding to the inputs the radiance field decoder ($\mathbf{p} \in \mathbb{R}^3$) and concatenate the conditioning variable $\mathbf{w}_{xyz}$ to the output of every other layer in the MLP starting from the input layer (\eg layers 0, 2, 4, and 6). To improve efficiency, we render a small resolution feature map of $512$ channels (two times smaller than the output resolution) instead of an RGB image and use a UNet \cite{unet} with additional deconvolution layers to predict the final image \cite{gsn, giraffe}. Fig. \ref{fig:model_architectures}(b) shows the architecture of the radiance field decoder module.
\end{itemize}

For training we initialize all latents $\mathbf{z} = 0$ and train them jointly with the parameters of the 3 modules. We use the Adam optimizer and a learning rate of $0.001$ for latents and $0.0001$ for model parameters. We train our model on 8 A100 NVIDIA GPUs for 2-7 days (depending on dataset size), with a batch size of $16$ trajectories where we randomly sample $2$ images per trajectory.

\begin{figure}
    \centering
    \begin{tabular}{cc}
    \includegraphics[width=0.5\textwidth]{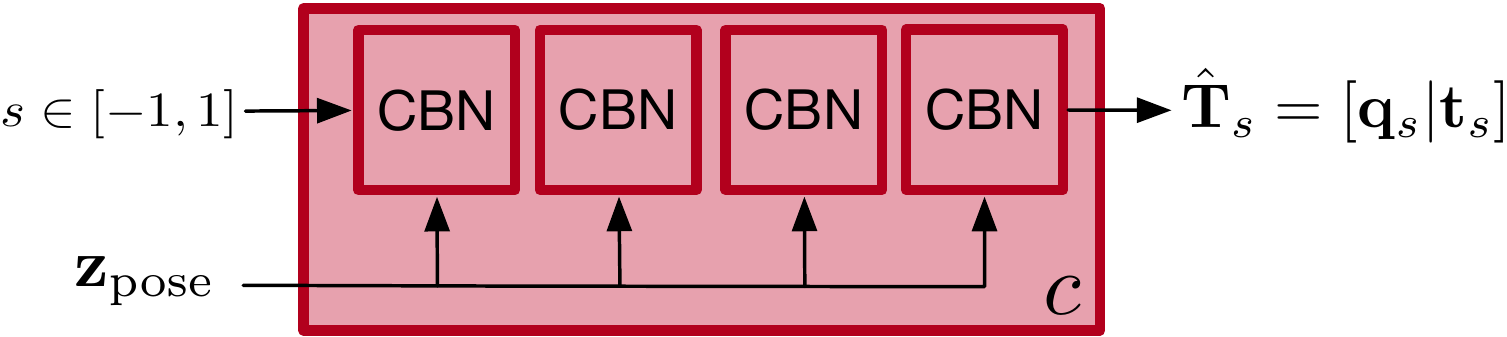} & \includegraphics[width=0.5\textwidth]{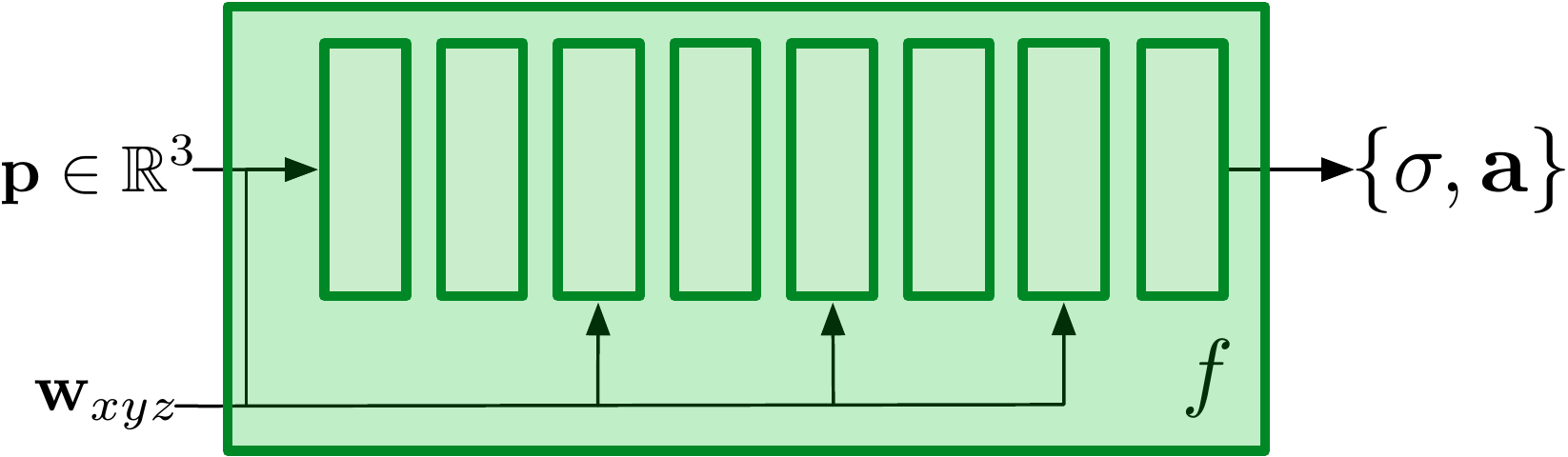} \\ 
    (a) & (b)

    \end{tabular}
    \caption{(a) Architecture of the camera pose decoder network. (b) Architecture of the radiance field network.}
    \label{fig:model_architectures}
\end{figure}

\section{Prior Architecture Design and Details}
~\label{app:priordetals}
We employ a Denoising Diffusion Probabilistic Model (DDPM) \cite{ddpm} to learn the distribution $p(Z)$. Specifically, we adopt the UNet architecture from \cite{improvedddpm} to denoise the latent at each timestep. During training, we sample $t \in \{1, ..., T\}$ uniformly and take the gradient descent step on $\theta_p$ from Eq. \ref{eq:ddpm_simple}. Different from \cite{improvedddpm}, we keep the original DDPM training scheme with fixed time-dependent covariance matrix and linear noise schedule. During inference, we start from sampling latent from zero-mean unit-variance Gaussian distribution and perform the denoising step iteratively. To accelerate the sampling efficiency, we leverage DDIM \cite{ddim} to denoise only 50 steps by modeling the deterministic non-Markovian diffusion processes.

For conditional generative modelling tasks, the conditioning mechanism should be general to support conditioning inputs from diverse modalities (i.e. text, image, categorical class, etc.). To fulfill this requirement, we first project the conditional inputs into an embedding representations $\mathbf{c}$ via a modality-specific encoder. For text conditioning, we employ a pre-trained RoBERTa-base \cite{roberta}. For image conditioning, we employ a ResNet-18 \cite{resnet} pre-trained on ImageNet. For categorical conditioning, we employ a trainable per-environment embedding layer. We freeze the encoders for text and image inputs to avoid over-fitting issues. We borrow the cross attention module from LDM \cite{ldm} to fuse the conditioning representation $\mathbf{c}$ with the intermediate activations at multiple levels in the UNet \cite{unet}. The cross-attention module implements an attention mechanism with key and value generated from $\mathbf{c}$ while the query generated from the intermediate activations in the UNet architecture (we refer readers to \cite{ldm} for more details).

For training the DDPM prior, we use the Adam optimizer and learning rate of $4.0e^{-06}$. We train our model on 1 A100 NVIDIA GPU for 1-3 days for unconditional prior learning and 3-5 days for conditional prior learning experiments (depending on dataset size), with a batch size of 256 and 32 respectively. For the hyper-parameters of the DDPM model, we set the number diffusion steps to 1000, noise schedule as linearly decreasing from 0.0195 to 0.0015, base channel size to 224, attention resolutions at [8, 4, 2, 1], and number of attention heads to 8.

\section{Ablation Study}
~\label{app:ablation}

We now provide additional ablations studies for the critical components in GAUDI. First, we analyze how the dimensionality of the latent code $z_d$ and the magnitude of $\beta$ affect the optimization problem defined in Eq. \ref{eq:autodecoder}. Tab. \ref{tab:ablation_reconstruction} shows reconstruction metrics for both RGB images and camera poses for a subset of $100$ trajectories in the VLN-CE dataset \cite{vlnce}. We observe a clear trend where increasing the magnitude of $\beta$ makes it harder to find latent codes with high reconstruction accuracy. This drop in accuracy is expected since $\beta$ controls the amount of noise in latent codes during training.  Finally, we observe that reconstruction performance starts to degrade when the latent code dimensionality grows past $2048$.

\begin{table}[h]
\scriptsize
    \begin{center}
    \begin{tabular}{c c c c c | c c}
    \toprule
           &  & $l_1$ $\downarrow$ & PSNR $\uparrow$ & SSIM $\uparrow$ & Rot Err. $\downarrow$ & Trans. Err $\downarrow$ \\ 
    \midrule
        $\beta=0.1$ & $z_d=2048$ & 7.63e-3 & 39.12 & 0.984 & 4.61e-3 & 2.90e-3\\ 
        $\beta=0.1$	& $z_d=4096$  & 7.89e-3 & 38.55 & 0.982 & 4.91e-3 & 2.76e-3\\ 
        $\beta=0.1$	& $z_d=8192$ & 9.02e-3 & 36.33 & 0.978 & 5.62e-3 & 3.36e-3\\ 
        $\beta=1.0$	& $z_d=2048$ & 1.00e-2 & 34.82 & 0.972 & 6.32e-3 & 3.77e-3\\ 
        $\beta=1.0$	& $z_d=4096$ & 1.11e-2 & 34.46 & 0.965 & 7.27e-3 & 5.69e-3\\ 
        $\beta=1.0$	& $z_d=8192$ & 1.54e-2 & 32.28 & 0.916 & 1.11e-2 & 7.13e-3\\ 
        $\beta=10.0$ & $z_d=8192$ & 3.89e-2 & 24.89 & 0.799 & 7.59e-2 & 3.61e-2\\ 
        $\beta=10.0$ & $z_d=4096$ & 9.25e-2 & 17.52 & 0.499 & 1.56e-1 & 6.30e-2\\ 
        $\beta=10.0$ & $z_d=2048$ & 1.35e-1 & 12.74 & 0.275 & 5.25e-1 & 1.29e-1\\ 
          
    \bottomrule
    \end{tabular}
    \end{center}
    \caption{Ablation experiment for the critical parameters of the optimization process described in Eq. \ref{eq:autodecoder}}
    \label{tab:ablation_reconstruction}
\end{table}

In addition, we also provide ablation experiments for the second stage of our model where we learn the prior $p(Z)$. In particular, we ablate critical factors of our model: the importance of learning corresponding scene and pose latents, the width of the denoising network in the DDPM prior, and the noise scale parameter $\beta$. In Tab. \ref{tab:ablation_random_pose_capacity} we show results for each factor. In particular, in the first two rows of Tab. \ref{tab:ablation_random_pose_capacity} we show the result of training the prior while breaking the correspondence of $\mathbf{z}=[\mathbf{z}_{\mathrm{pose}}, \mathbf{z}_{\mathrm{scene}}]$.  We break this correspondence by forming random pairs of $\mathbf{z}=[\mathbf{z}_{\mathrm{pose}}, \mathbf{z}_{\mathrm{scene}}]$ after optimizing the latent representations, and then training the prior on these random pairs. We observe that training the prior to render scenes from a random pose latent impacts both the FID and SwAV-FID scores substantially, which provides support for our claim that the distribution of valid camera poses depends on the scene. In addition, we can see how the width of the denoising model affects performance.  By increasing the number of channels, the DDPM prior is able to better capture the distribution of latents. Finally, we also show how different noise scales $\beta$ impact the capacity of the generative model to capture the distribution of scenes. All results Tab \ref{tab:ablation_random_pose_capacity} are performed on the full VLN-CE dataset \cite{vlnce}.

\begin{table}[h]
\scriptsize
\begin{center}
\setlength\tabcolsep{2pt} 
 \begin{tabular}{lcc}
    \toprule
    &  \multicolumn{2}{c}{VLN-CE \cite{vlnce}} \\
   \cmidrule(r){2-3}
  & FID $\downarrow$ & SwAV-FID $\downarrow$ \\ 
 \midrule 
  GAUDI &  $18.52$ &  $3.63$ \\
  GAUDI w. Random Pose &  $83.66$ &  $10.73$ \\
 \midrule 
  Base Channel Size = 64 &  $104.27$ &  $13.21$ \\
  Base Channel Size = 128 &  $22.04$ &  $4.35$ \\
  Base Channel Size = 192 &  $18.61$ &  $3.79$ \\
  Base Channel Size = 224 &  $18.52$ &  $3.63$ \\
 \midrule 
  Noise Scale $\beta$ = 0.0 &  $18.48$ &  $3.68$ \\
  Noise Scale $\beta$ = 0.1 (same as 1st stage) &  $18.52$ &  $3.63$ \\
  Noise Scale $\beta$ = 0.2 &  $18.48$ &  $3.67$ \\
  Noise Scale $\beta$ = 0.5 &  $20.20$ &  $4.11$ \\
  \bottomrule
\end{tabular} 
\end{center}
\caption{Ablation study for different design choice of GAUDI. }
\label{tab:ablation_random_pose_capacity}
\end{table}

In Tab. \ref{tab:ablation_conditioning_mechanism} we report ablation results for modulation on the denoising architecture in the DDPM prior. We compare cross-attention style conditioning as in LDM \cite{ldm} with FiLM style conditioning \cite{film}.
For FiLM style conditioning, we take the mean of the conditioning representation $\mathbf{c}$ across spatial dimension and project it into the same space as denoising timestep embedding. After that, we take the sum of the conditioning and timestep embedding and predict the scaling and shift factors of the affine transformation applied to the UNet intermediate activations. We compare the performance of the two conditioning mechanisms in Tab.  \ref{tab:ablation_conditioning_mechanism}. We observe that the cross-attention style conditioning  performs better than the FiLM style across all our conditional generative modeling experiments.

\begin{table}[h]
\scriptsize
\begin{center}
% \resizebox{\columnwidth}{!}{%
\setlength\tabcolsep{2pt} % default value: 6pt
 \begin{tabular}{lcccccc}
    \toprule
    &  \multicolumn{2}{c}{Text Conditioning} & \multicolumn{2}{c}{Image Conditioning} &  \multicolumn{2}{c}{Categorical Conditioning} \\
   \cmidrule(r){2-3}\cmidrule(r){4-5}\cmidrule(r){6-7}
  & FID $\downarrow$ & SwAV-FID $\downarrow$ & FID $\downarrow$ & SwAV-FID $\downarrow$ & FID $\downarrow$ & SwAV-FID $\downarrow$\\ 
 \midrule 
  FiLM Module \cite{film} &  $20.99$ &  $4.11$  & $ 21.01 $ & $ 4.21 $ & $18.75$ &  $3.63$ \\
  Cross Attention \cite{ldm} &  $18.50$ &  $3.75$ & $ 19.51 $ & $ 3.93 $ &  $18.74$ &  $3.61$ \\
  \bottomrule
\end{tabular} 
\end{center}
\caption{Ablation study for conditioning mechanism of GAUDI.}
\label{tab:ablation_conditioning_mechanism}
\end{table}

\subsection{Additional Visualizations}
~\label{app:visualizations}

\begin{figure}[!h]
    \centering
    \includegraphics[width=\textwidth]{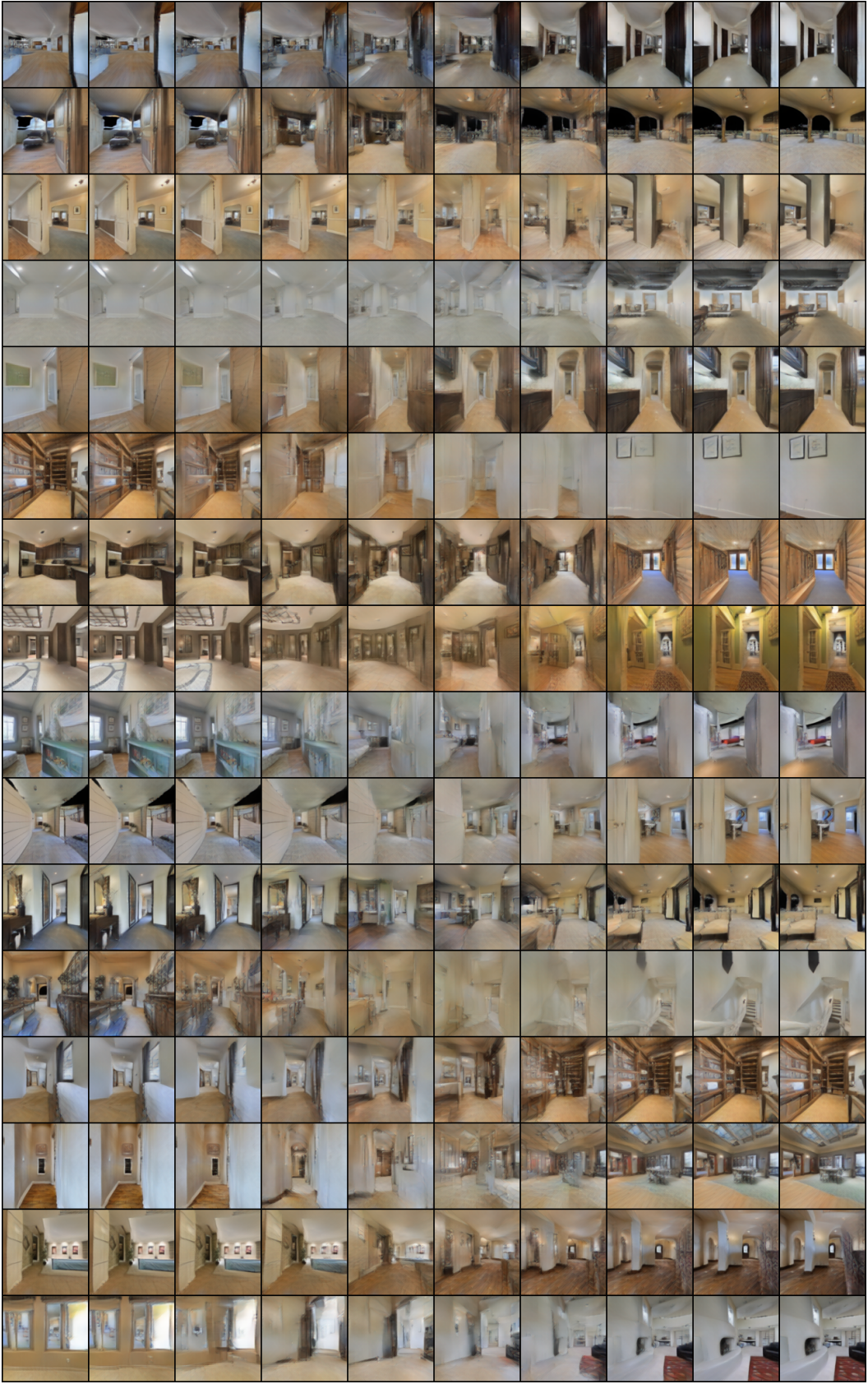}
    \caption{Additional interpolation of 3D scenes in latent space for the VLN-CE dataset \cite{vlnce}. Each row corresponds to a different interpolation path between random pairs of latent representations $(\mathbf{z}_i, \mathbf{z}_j)$.}
    \label{fig:app_interp_random}
\end{figure}

\begin{figure}[!h]
    \centering
    \includegraphics[width=\textwidth]{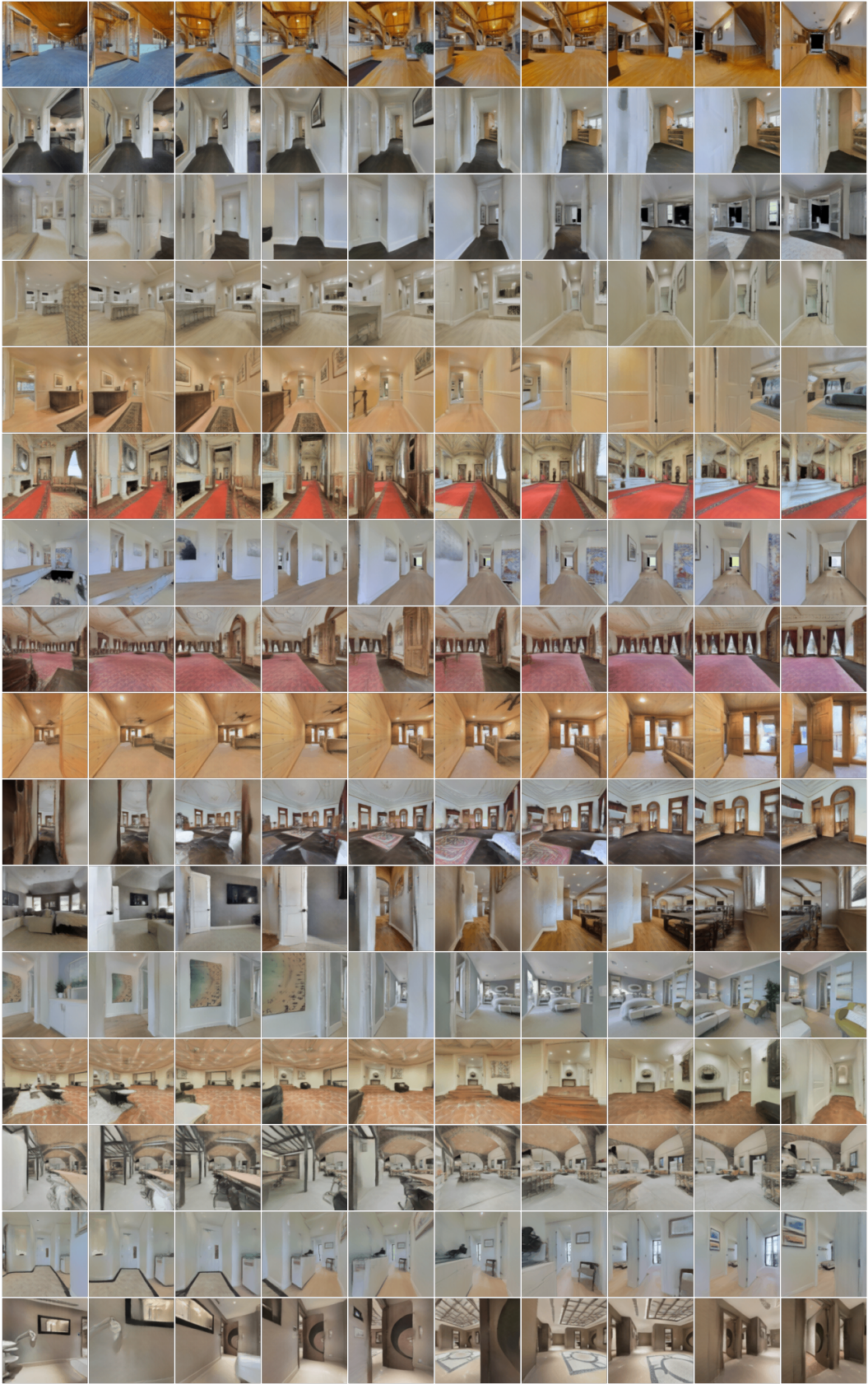}
    \caption{Additional visualizations of scenes sampled from unconditional GAUDI for VLN-CE dataset \cite{vlnce}. Each row to a scene rendered from camera poses sampled from the prior. }
    \label{fig:app_uncond_vlnce}
\end{figure}

\begin{figure}[!h]
    \centering
    \includegraphics[width=\textwidth]{figs/appendix/uncond_arkit.pdf}
    \caption{Additional visualizations of scenes sampled from unconditional GAUDI for ARKitScenes dataset \cite{arkit}. Each row corresponds to a scene rendered from camera poses sampled from the prior. }
    \label{fig:app_uncond_arkit}
\end{figure}

\begin{figure}[!h]
    \centering
    \includegraphics[width=\textwidth]{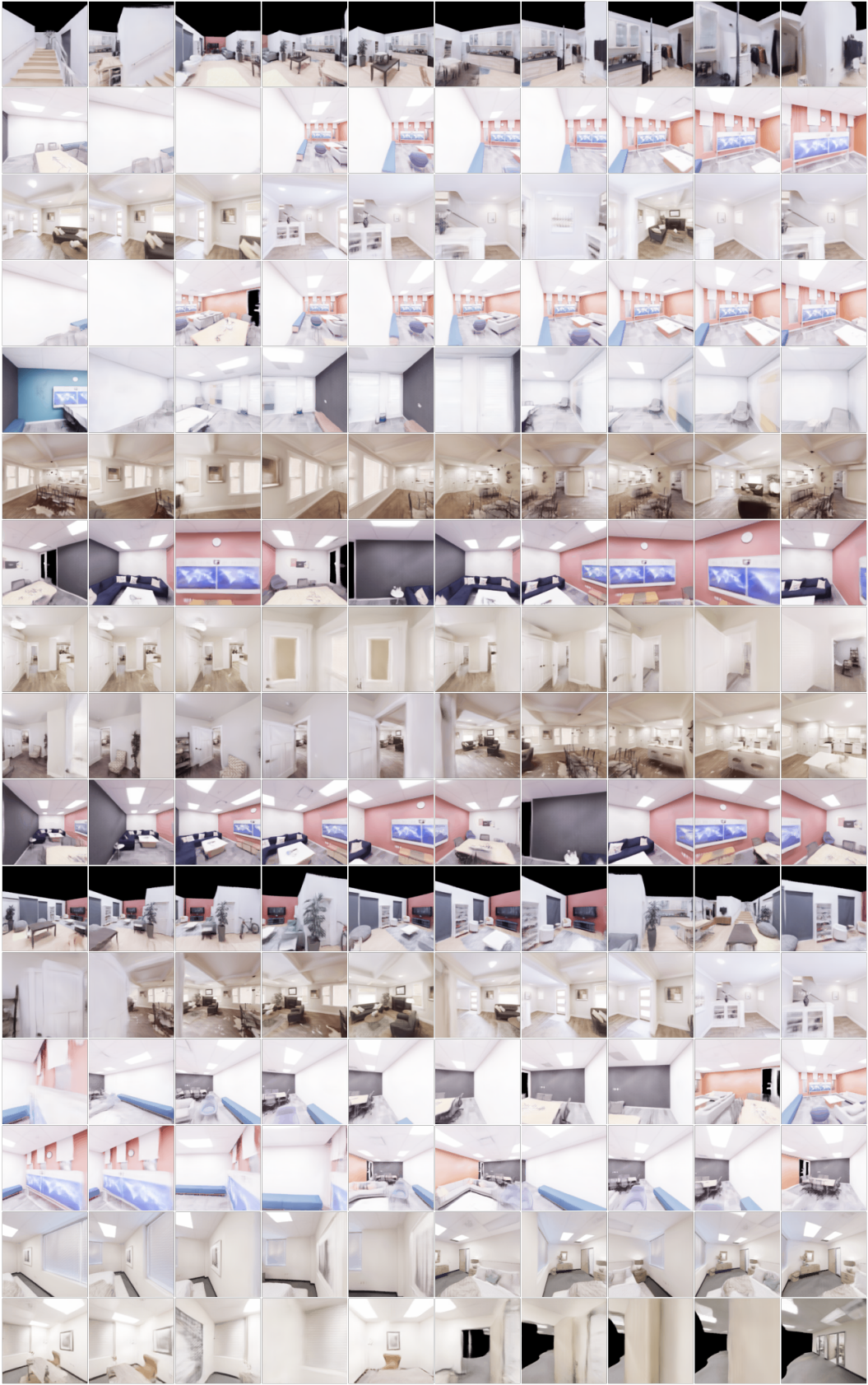}
    \caption{Additional visualizations of scenes sampled from unconditional GAUDI for Replica dataset \cite{replica}. Each row corresponds to a scene rendered from camera poses sampled from the prior. }
    \label{fig:app_uncond_replica}
\end{figure}

In this section we provide additional visualizations for both figures in this appendix and videos that can be found attached in the supplementary material. In Fig. \ref{fig:app_interp_random} we provide additional interpolations between random pairs of latents obtained for VLN-CE dataset \cite{vlnce}, where each row represents a interpolation path between a random pair of latents (\ie rightmost and leftmost columns). We can see how the model tends to produce smoothly changing interpolation paths which align similar scene content. In addition we refer readers to the folder \url{./interpolations} in which videos of interpolations can be found where for each interpolated scene we immersively navigate it by moving the camera forwards and rotating left and right.

In addition, we provide more visualization of samples from the unconditional GAUDI model in Fig. \ref{fig:app_uncond_vlnce} for VLN-CE \cite{vlnce}, Fig. \ref{fig:app_uncond_arkit} for ARKitScenes \cite{arkit} and Fig. \ref{fig:app_uncond_replica} for Replica \cite{replica}. In all these figures, each row represents a sample from the prior that is rendered from its corresponding sampled camera path. We note how these qualitative results reinforce the fidelity and variability of the distribution captured by GAUDI, which is also reflected in the quantitative results in Tab. \ref{tab:generative_quant} of the main paper. In addition, the folder \url{./uncond_samples} contains videos of more samples from the unconditional GAUDI model for all datasets.

Finally, the folder \url{./cond_samples} contains a video showing samples from GAUDI conditioned on different modalities like text, images or categorical variables. These visualizations corresponds to the results in Sect.  \ref{sec:conditional_experiments} of the main paper.

\section{License}
Due to licensing issues we cannot release the VLN-CE \cite{vlnce} raw trajectory data and we refer the reader to \url{https://github.com/jacobkrantz/VLN-CE} and to the license of the Matterport3D data \url{http://kaldir.vc.in.tum.de/matterport/MP_TOS.pdf}.

\end{document}